\documentclass[letterpaper]{article} 
\usepackage{aaai2026}  
\usepackage{times}  
\usepackage{helvet}  
\usepackage{courier}  
\usepackage[hyphens]{url}  
\usepackage{graphicx} 
\usepackage{natbib}  
\usepackage{booktabs}
\usepackage{caption} 
\usepackage{amsmath, amssymb}
\usepackage{multirow}
\usepackage{algorithm}
\usepackage{algorithmic}

\usepackage{newfloat}
\usepackage{listings}
\DeclareCaptionStyle{ruled}{labelfont=normalfont,labelsep=colon,strut=off} 
\floatstyle{ruled}
\newfloat{listing}{tb}{lst}{}
\floatname{listing}{Listing}
\title{EMOD: A Unified EEG Emotion Representation Framework Leveraging V-A Guided Contrastive Learning}
\author {
	Yuning Chen\textsuperscript{\rm 1,2},
	Sha Zhao\textsuperscript{\rm 1,2}\thanks{Corresponding authors},
	Shijian Li\textsuperscript{\rm 1,2},
	Gang Pan\textsuperscript{\rm 1,2,3},
}

\affiliations {
	\textsuperscript{\rm 1}State Key Laboratory of Brain-machine Intelligence, Zhejiang University\\
	\textsuperscript{\rm 2}College of Computer Science and Technology, Zhejiang University\\
	\textsuperscript{\rm 3}MOE Frontier Science Center for Brain Science and Brain-machine Integration,
	Zhejiang University\\
	\{yuningchen,  szhao, shijianli, gpan\}@zju.edu.cn
}

\usepackage{bibentry}

\begin{document}

\maketitle

\begin{abstract}
	Emotion recognition from EEG signals is essential for affective computing and has been widely explored using deep learning. While recent deep learning approaches have achieved strong performance on single EEG emotion datasets, their generalization across datasets remains limited due to the heterogeneity in annotation schemes and data formats. Existing models typically require dataset-specific architectures tailored to input structure and lack semantic alignment across diverse emotion labels.
	To address these challenges, we propose EMOD: A Unified EEG Emotion Representation Framework Leveraging Valence-Arousal (V-A) Guided Contrastive Learning. EMOD learns transferable and emotion-aware representations from heterogeneous datasets by bridging both semantic and structural gaps. Specifically, we project discrete and continuous emotion labels into a unified V-A space and formulate a soft-weighted supervised contrastive loss that encourages emotionally similar samples to cluster in the latent space.
	To accommodate variable EEG formats, EMOD employs a flexible backbone comprising a Triple-Domain Encoder followed by a Spatial-Temporal Transformer, enabling robust extraction and integration of temporal, spectral, and spatial features. We pretrain EMOD on 8 public EEG datasets and evaluate its performance on three benchmark datasets. Experimental results show that EMOD achieves the state-of-the-art performance, demonstrating strong adaptability and generalization across diverse EEG-based emotion recognition scenarios.
\end{abstract}

\begin{links}
    \link{Code}{https://github.com/cyn4396/EMOD}
\end{links}

\begin{figure}[t]
	\centering
	\includegraphics[width=0.48\textwidth]{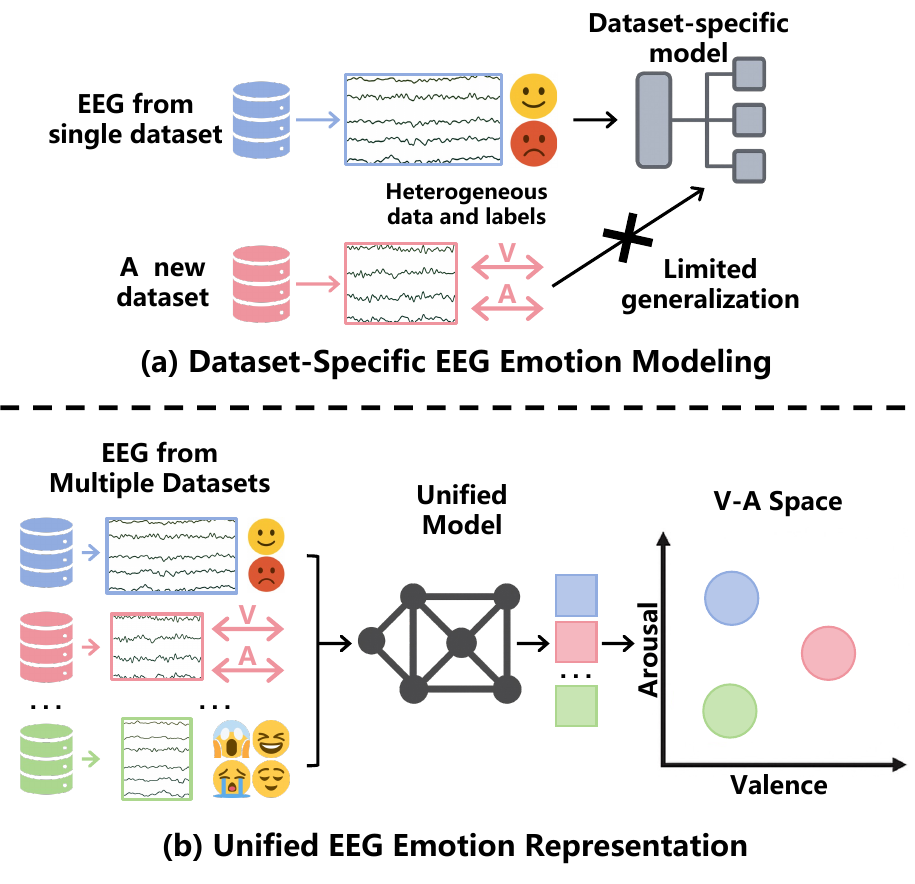}
	\caption{Comparison between conventional EEG emotion recognition and our EMOD framework. Traditional methods rely on dataset-specific supervision and manually designed architectures, limiting generalization across datasets. In contrast, EMOD employs V-A guided pretraining on heterogeneous data, enabling robust and generalizable emotion-aware representations.}
	\label{fig_compare}
\end{figure}

\section{Introduction}
Emotion is a fundamental component of human life, influencing cognition, behavior, and social interaction~\cite{lench_functional_2013}. The ability to understand and recognize emotions is crucial for the development of affective computing—a field focuses on bridging the gap between human emotions and intelligent systems~\cite{wang_systematic_2022}. By enabling more natural and adaptive human-computer interactions, affective computing highlights the growing need for accurate, robust, and scalable emotion recognition methods~\cite{mathur_2023}. Among various sensing modalities, electroencephalography (EEG) stands out as a promising tool for emotion recognition. Unlike behavioral signals—such as facial expressions, vocal tone, or body gestures—that are often influenced by cultural norms or conscious control, EEG offers a direct and objective measure of neural activity underlying emotional processes. Its high temporal resolution and capacity to capture fine-grained brain dynamics make EEG well-suited for decoding affective states with improved robustness and precision.

Recently, numerous deep learning models have been proposed for EEG-based emotion recognition~\cite{song_eeg_2020, ding_tsception_2023, zhao2025music2emotion}. However, as illustrated in Figure~\ref{fig_compare}(a), these models are typically designed for specific datasets, making joint training and deployment across different datasets difficult. Their generalizability are limited in practical applications where there are usually new EEG sources appearing. A major obstacle to such generalization lies in the inherent heterogeneity of EEG datasets, which manifests in two fundamental aspects. 
First,~\textbf{label heterogeneity} arises from differences in annotation schemes. Some datasets adopt discrete emotion labels (e.g., happy, sad, fear) often derived from video stimuli or expert rating, while others rely on continuous self-reported valence-arousal (V-A) scores. Even within the same paradigm, label definitions and value ranges often vary, resulting in semantic mismatches that hinder model transferability across datasets.
Second,~\textbf{data format heterogeneity} arises from variations in EEG acquisition protocols, such as different channel configurations and sampling rates. For instance, some datasets use 62-channel EEG at 200 Hz, while others employ 32-channel signals sampled at 128 HZ. These discrepancies force prior models to adopt dataset-specific architectures, limiting their scalability and preventing unified modeling across diverse EEG formats.

To address these challenges, we propose EMOD, a pretraining framework designed to learn generalizable and emotion-aware EEG representations from heterogeneous datasets, as illustrated in Figure~\ref{fig_compare}(b). EMOD aims to bridge both the semantic and structural gaps across datasets, enabling unified, scalable, and transferable representation learning for EEG-based emotion recognition.
\textbf{To address label heterogeneity}, we leverage the widely accepted Valence-Arousal (V-A) emotion model~\cite{russell1980circumplex} to \textbf{project all available emotion annotations—whether discrete or continuous—into a shared affective space}. Based on this projection, we design a soft-weighted supervised contrastive loss that defines inter-sample similarity by V-A proximity, encouraging representations that reflect the continuous emotional semantics across datasets. To our best knowledge, there have been no studies considering the emotion label consistency for the EEG emotion recognition task, and we are the first aiming to address this limitation.
\textbf{To handle data format heterogeneity}, we devise a flexible and unified architecture that supports varying channel configurations, sampling rates, and segment lengths. It combines a Triple-Domain Encoder and Spatial-Temporal Transformer to extract temporal, spectral and spatial features across diverse EEG formats.
Together, EMOD tries to learn semantically meaningful and structurally adaptive EEG representations, facilitating more robust and generalizable emotion recognition across heterogeneous datasets.
Our main contributions are summarized as follows:
\begin{itemize}
	\item We propose EMOD, a unified pretraining framework for EEG-based emotion recognition that leverages V-A guided contrastive learning to learn generalizable and emotion-aware representations from heterogeneous datasets. EMOD is pretrained on 8 public EEG datasets and evaluated on 3 widely-used EEG emotion benchmarks. Extensive experiments demonstrate that EMOD achieves state-of-the-art performance and strong generalization across diverse datasets.
	\item For the first challenge, we devise a V-A guided contrastive learning strategy that unifies discrete and continuous emotion labels into a shared affective space, enabling semantic alignment and improving label compatibility across datasets. To our best knowledge, we are the first to address the inconsistency of EEG emotion labels across different datasets.
	\item For the second challenge, we design a flexible backbone with a Triple-Domain Encoder and Spatial-Temporal Transformer to capture temporal, spectral, and spatial features across diverse EEG formats.
\end{itemize}

\section{Related Works}
\subsection{EEG-Based Emotion Recognition}
Traditional EEG-based emotion recognition relies on handcrafted features and subsequent classification~\cite{psd_2015,duan2013differential,s19071631}, but struggles to capture the nonlinear and dynamic nature of EEG signals. Recent works address this by adopting end-to-end deep learning models that learn hierarchical features directly from raw data. A variety of architectures have been explored, including CNNs~\cite{liu_3dcann_2022}, RNNs~\cite{tao_eeg-based_2023}, GNNs~\cite{zhong_eeg-based_2022}, and Transformer~\cite{ding_emt_2024} hybrids. 
Beyond architecture, contrastive learning improves transferability by enforcing semantic similarity~\cite{shen_contrastive_2023, li_generalized_2024, li_gmss_2023}, while continual learning addresses EEG non-stationarity and enhances adaptability~\cite{Zhou2025BrainUICLAU, zhou2025spiced}.
Despite good performance, existing approaches face two limitations: (1) they are largely restricted to single-dataset training and lack generalizability across datasets with diverse formats and annotation schemes; (2) they rely on discrete emotion labels and assume uniform similarity within classes, overlooking the continuous and nuanced structure of human emotions.
\begin{figure*}[t]
	\centering
	\includegraphics[width=1\textwidth]{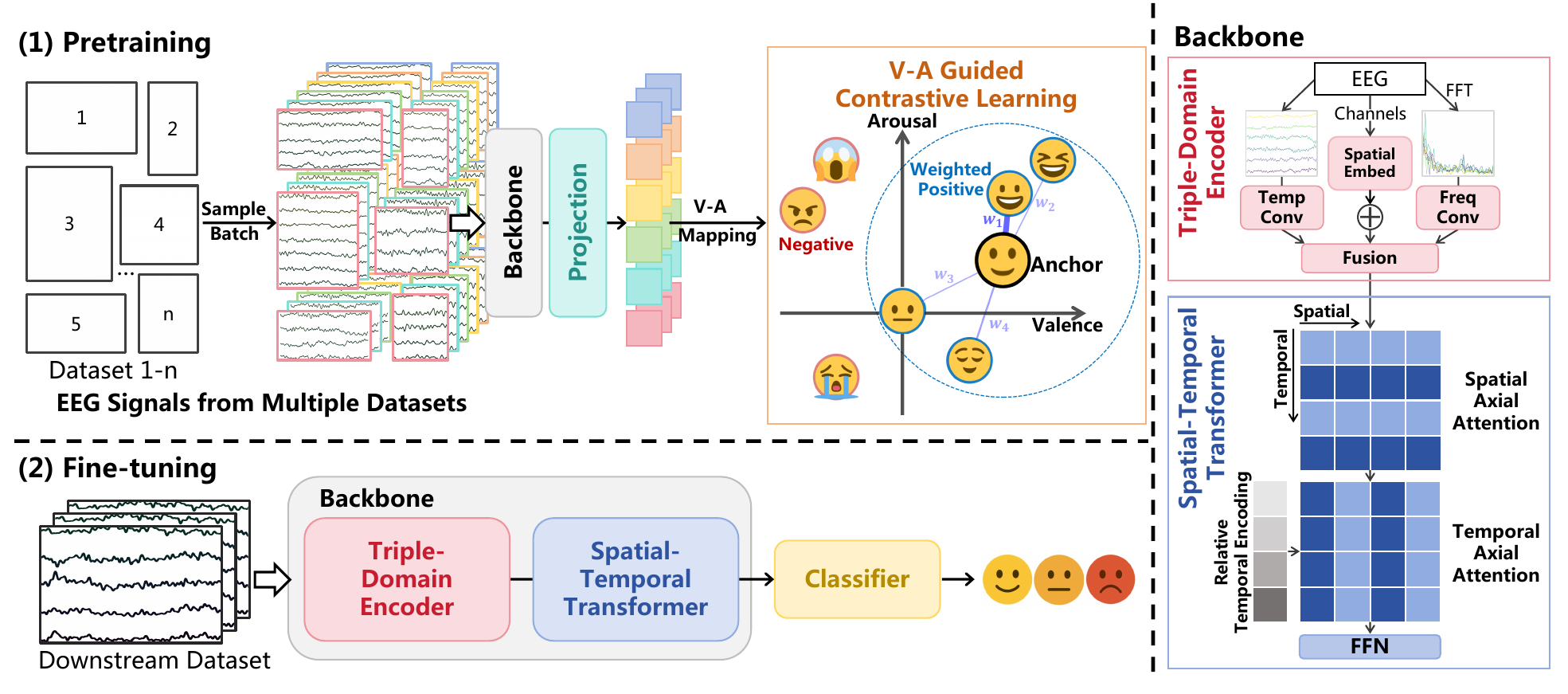}
	\caption{Overview of the EMOD framework.}
	\label{fig0}
\end{figure*}

\subsection{Representation Learning for EEG}
Recent advances in self-supervised learning have enabled large-scale EEG representation pretraining. Several foundation models have emerged, including Brant~\cite{zhang2023brant} with masked modeling for neural signals, LaBraM~\cite{jiang2024large} with channel-wise spectrum prediction, CBraMod~\cite{wang2024cbramod} using criss-cross attention, and EEGMamba~\cite{WANG2025107816} based on a Mamba state-space model.
Although these models show strong transferability in general EEG tasks, they are usually optimized for low-level objectives such as reconstruction or masking, which capture structural patterns but overlook emotion-relevant dynamics. As a result, the learned representations lack emotion-specific inductive biases, limiting their effectiveness for downstream affective tasks.

\section{Method}

Recent advances in self-supervised learning have enabled effective representation learning from large-scale unlabeled EEG data. Several foundation models have been proposed to extract generalized features across tasks and datasets. For instance, Brant~\cite{zhang2023brant} adopts masked modeling for intracranial neural signals; LaBraM~\cite{jiang2024large} performs channel-wise segmentation and vector-quantized spectrum prediction; CBraMod~\cite{wang2024cbramod} leverages criss-cross attention and asymmetric positional encoding to model spatial-temporal dependencies; and EEGMamba~\cite{WANG2025107816} introduces a Mamba-based state-space model for large-scale EEG pretraining.
While these models show strong transferability in general EEG applications, they face critical limitations in emotion recognition. Most foundation models are optimized for low-level objectives such as signal reconstruction or masked prediction, which capture generic EEG structures but overlook emotion-relevant dynamics. As a result, their learned representations lack emotion-specific inductive biases, reducing their effectiveness for affective tasks.

\subsection{Cross-Dataset Adaptive EEG Backbone Architecture}
To extract generalizable EEG representations from heterogeneous datasets, we design a flexible backbone that captures emotion-relevant features from three complementary domains. It comprises a Triple-Domain Encoder module, which integrates temporal, spectral, and spatial information, followed by a Spatial-Temporal Transformer that models long-range dependencies across both dimensions.

\subsubsection{Triple-Domain Encoder.}

To extract emotion-relevant features from temporal, spectral, and spatial domains, we design a triple-domain encoder module that integrates information into a coherent token representation, which is then fed into the spatial-temporal transformer to capture higher-level spatiotemporal dependencies and emotional dynamics.

Given an input EEG segment $\mathbf{X} \in \mathbb{R}^{C \times T}$, where $C$ is the number of channels and $T$ the number of time points, we first extract \textbf{temporal-domain tokens} $\mathbf{X}_{\text{t}} \in \mathbb{R}^{C \times T’ \times d}$ by directly processing raw signals, and \textbf{spectral-domain tokens} $\mathbf{X}_{\text{f}} \in \mathbb{R}^{C \times T’ \times d}$ via Fast Fourier Transform (FFT), where $T’$ is the number of temporal segments derived from convolution, and $d$ is the embedding dimension. These two types of tokens are concatenated along the embedding dimension and linearly projected to form the temporal-spectral representation:
\begin{equation}
	\mathbf{X}_{\text{tf}} = \text{Linear} \left( \left[ \mathbf{X}_{\text{t}}; \mathbf{X}_{\text{f}} \right] \right) \in \mathbb{R}^{C \times T’ \times d},
\end{equation}

To generalize across datasets with varying channel configurations, we introduce a learnable \textbf{channel embedding} matrix $\mathbf{E}_s \in \mathbb{R}^{C \times d}$, where each row vector $\mathbf{e}_i$ encodes the anatomical and identity information of the $i$-th electrode. These spatial embeddings are broadcast along the temporal axis and added to the fused tokens:
\begin{equation}
	\mathbf{H} = \mathbf{X}_{\text{tf}} + \mathbf{E}_s \in \mathbb{R}^{C \times T’ \times d},
\end{equation}

This triple-domain encoder captures complementary time-frequency patterns while remaining robust to diverse electrode configurations across datasets.

\subsubsection{Spatial-Temporal Transformer.}

To learn from heterogeneous EEG formats, we adopt an axial attention mechanism~\cite{ho2019axial} that separately captures spatial and temporal dependencies, enabling effective modeling of cross-channel and temporal interactions. Specifically, \textbf{spatial axial attention} is applied across EEG channels at each time step to capture inter-electrode relationships, while \textbf{temporal axial attention} models intra-channel dynamics over time, capturing long-range affective dependencies.

While spatial attention benefits from absolute positional encodings—since EEG channels correspond to anatomically fixed electrode locations—the same approach does not apply well to the temporal dimension. Time steps in EEG signals lack consistent semantic meaning. For instance, the emotional state at second 2 may not inherently carry more significance than at second 10. Using absolute temporal positions may therefore introduce bias or overfitting.
To address this, we incorporate a \textbf{relative temporal encoding} scheme into the temporal axial attention. Specifically, we compute a relative distance matrix $\mathbf{R} \in \mathbb{R}^{T’ \times T’}$ that encodes pairwise temporal offsets between tokens. This matrix is then added to the attention logits to modulate similarity based on relative rather than absolute time:

\begin{equation}
	\text{Attention}_{\text{tem}}(\mathbf{Q}, \mathbf{K}, \mathbf{V}) = \text{softmax}\left( \frac{\mathbf{QK}^\top}{\sqrt{d}} + \mathbf{R} \right) \mathbf{V},
\end{equation}

This design allows the model to capture temporal dependencies based on relative positions, enabling a more flexible representation of evolving emotional dynamics.

\subsection{V-A Guided Contrastive Learning}

A major obstacle to generalizable EEG-based emotion recognition lies in the inconsistency of emotion annotations: some datasets adopt discrete emotion categories (e.g., “happy”, “sad”), while others provide continuous V-A scores. This label heterogeneity hinders unified representation learning and cross-dataset alignment.
To address this, we try to construct a unified affective space based on V-A emotion model, into which all the labeled EEG segments are projected. This continuous space enables the comparison of emotional similarity across datasets regardless of their original label format.

\subsubsection{Constructing the Unified V-A Space and Cross-Dataset Sampling}

To enable semantically aligned representation learning across heterogeneous EEG emotion datasets, we construct a unified affective space grounded in the V-A emotion model. For datasets with V-A annotations, we directly use the provided coordinates. For those with only discrete emotion labels, we estimate V-A values by averaging statistical mappings from~\cite{cowen2017self}. All coordinates are linearly rescaled to the range [–4, 4] for consistency. To provide a structured emotional reference, we divide both valence and arousal axes into 9 integer levels from –4 to 4, forming a $9 \times 9$ grid where each cell corresponds to a distinct affective region. This design preserves the continuity of emotional semantics while supporting fine-grained, cross-dataset representation alignment.

We further group the $9 \times 9$ grid into $3 \times 3$ macro regions, resulting in 9 broader affective zones to promote balanced and semantically diverse batch construction. For each source dataset, we sample each mini-batch via $m$ rounds of region-wise sampling. In each round, one EEG segment is randomly drawn from each of the 9 macro regions in every dataset. We obtain $9 \times m$ samples for each dataset.  Given $n$ source datasets, such process results in a batch of size $n \times 9 \times m$, providing both dataset-level coverage and affective diversity.

This structured sampling strategy encourages consistent exposure to a wide range of emotional states and mitigates distributional biases across datasets. By aligning samples in a shared V-A space and enforcing semantic balance within each batch, the model is better equipped to learn smooth and generalizable emotional representations across domains.

\subsubsection{Distance-Aware Contrastive Loss}

A key component of contrastive learning is the definition of positive and negative sample pairs. Traditional methods—such as Supervised Contrastive Loss (SupCon)~\cite{khosla2020supervised}—assume discrete labels and treat all the samples within the same class as equally positive:

\begin{equation}
	\mathcal{L}^{\text{sup}} = \sum_{i \in I} \frac{-1}{|P(i)|} \sum_{p \in P(i)} \log \frac{\exp(z_i^\top z_p / \tau)}{\sum_{a \in A(i)} \exp(z_i^\top z_a / \tau)},
\end{equation}
where $z_i$ and $z_p$ are normalized embeddings of the anchor and positive sample, $\tau$ is a temperature scaling parameter, $I$ is the set of all the anchors in the mini-batch, $P(i)$ denotes the set of positives (same class as anchor), and $A(i)$ includes all samples except $i$.

However, such binary treatment of similarity overlooks the continuous nature of emotions. In real-world emotion datasets, emotions often exhibit graded semantic relationships—for instance, “happy” is emotionally closer to “excited” than to “sad”, even if all the three belong to distinct classes. Rigidly treating all non-identical classes as negatives may suppress this valuable structure.

To better reflect the graded nature of emotion, we propose \textbf{a distance-aware contrastive loss with soft weights based on emotional proximity in the shared V-A space}. For each anchor $i$, all the other samples $j$ within an emotional radius $d_{\text{max}}$ are treated as positives, but rather than equally weighted, they are assigned \textbf{soft similarity weights}:

\begin{equation}
	w_{ij} = \max\left( 0, 1 - \frac{d_{ij}}{d_{\text{max}}} \right),
\end{equation}
where $d_{ij}$ is the Euclidean distance between samples $i$ and $j$ in the V-A space. Pairs farther than $d_{\text{max}}$ are assigned zero weight and treated as negatives. These soft weights allow the model to consider the continuous emotional similarity between samples, instead of relying on binary class equivalence (i.e., 0 or 1).
The distance-aware contrastive loss is then defined as:

\begin{equation}
	\mathcal{L}_{\text{V-A}} =\sum_{i \in I} \frac{-1} {\sum\limits_{j \ne i}  {w_{ij}}} \sum_{j \ne i} w_{ij} \cdot \log \frac{\exp\left( z_i^\intercal z_j / \tau \right)}{
		\sum\limits_{k \ne i} \exp\left( z_i^\intercal z_k / \tau \right)},
	\label{eq:va_soft_contrastive}
\end{equation}

By integrating soft weights into the contrastive objective, our approach captures nuanced affective relationships: emotionally similar samples contribute more to the loss, while distant ones are down-weighted or ignored. Embedding emotional semantics directly into the pretraining objective encourages the model to learn a semantically structured representation space, which offers several key benefits:
(1) \textbf{Modeling intra-class variability} — capturing fine-grained affective differences within a single category (e.g., subtle shades of “happy”);
(2) \textbf{Preserving inter-class continuity} — encoding smooth emotional transitions across categories (e.g., gradients from “calm” to “excited”);
(3) \textbf{Enabling multi-dataset alignment} — mapping heterogeneous emotion labels into a unified V-A space for consistent cross-dataset representation.
Collectively, these capabilities lead to more robust, expressive, and generalizable EEG encodings for downstream emotion recognition tasks.

After V-A guided contrastive pretraining, we perform \textbf{fine-tuning} on target EEG emotion recognition tasks using labeled data. A simple linear classifier is appended to the pretrained encoder, and the entire model is optimized with cross-entropy loss.

\begin{table}[!t]
	\centering
	\setlength{\tabcolsep}{2.5pt}
	\small
	\begin{tabular}{lcccc}
		\toprule
		\textbf{Dataset} & \textbf{Sampling Rate} & \textbf{Channels} & \textbf{Subjects} & \textbf{Samples} \\
		\midrule
		DEAP            & 128                         & 32        &    32     & 3840            \\
		SEED-IV          & 200                         & 62       &   15       & 14655            \\
		SEED-VII          & 200                         & 62        &   20       &27340             \\
		AMIGOS           & 128                         & 14          &   40    & 4389            \\
		MPED            & 200                         & 28            &  23   & 12190            \\
		DREAMER            & 128                         & 14       &   23      & 8372             \\
		MAHNOB-HCI & 256                        & 32         &  27    & 3955             \\
		MDME & 300                         & 21         &    73    & 5329\\
		\bottomrule
	\end{tabular}
	\caption{Statistics of the EEG datasets used for pretraining}
	\label{tab:pretrain_dataset}
\end{table}

\begin{table*}[!t]
	\centering
	\resizebox{\textwidth}{!}{%
		\begin{tabular}{lcccccccccc}
			\toprule
			& & \multicolumn{3}{c}{\textbf{FACED (9 classes, 10s)}} & \multicolumn{3}{c}{\textbf{SEED-V (5 classes, 1s)}} & \multicolumn{3}{c}{\textbf{SEED (3 classes, 5s)}} \\
			\cmidrule(lr){3-5} \cmidrule(lr){6-8} \cmidrule(lr){9-11}
			\textbf{Method} &\textbf{Params}  & \textbf{BACC} & \textbf{Kappa} & \textbf{WF1} 
			& \textbf{BACC} & \textbf{Kappa} & \textbf{WF1} 
			& \textbf{BACC} & \textbf{Kappa} & \textbf{WF1} \\
			\midrule
			EEGNet & 0.003M       & 40.90/1.22 & 33.42/2.51 & 41.24/1.41 & 29.61/1.02 & 10.06/1.43 & 27.49/0.98 & 51.11/1.44 & 26.79/2.20 & 51.05/2.26 \\
			EEGConformer&0.55M   & 45.59/1.25 & 38.58/1.86 & 45.14/1.07 & 35.37/1.12 & 17.72/1.74 & 34.87/1.36 &41.54/1.31 & 12.27/1.99& 41.27/1.47 \\
			ContraWR&1.6M       & 48.87/0.78 & 42.31/1.51 & 48.84/0.74 & 35.46/1.05 & 19.05/1.88 & 35.44/1.21 & 57.96/0.74 & 37.17/1.15 & 58.46/0.78 \\
			ST-Transformer&3.5M  & 48.10/0.79 & 41.37/1.33 & 47.95/0.96 & 30.52/0.72 & 10.83/1.21 & 28.33/1.05 & 48.27/1.96 & 22.50/3.01 & 48.07/2.31 \\
			Tsception&0.03M  & 18.40/1.80 & 8.35/2.05 & 11.54/3.01 & 38.52/1.37 & 23.39/1.63 & 38.98/1.51 & 54.24/1.13 & 31.62/1.70 & 54.60/1.39 \\
			Icaps-ReLSTM &1.1M  & 43.74/0.41 & 36.48/0.45 & 43.69/0.42 & 35.61/0.56 & 19.46/0.78& 36.10/0.69& 53.67/0.39 & 30.74/0.56 & 53.93/0.46 \\
			MACTN&6.8M          & 43.34/2.83 & 36.00/3.27 & 43.87/2.90 & 37.28/0.90 & 21.61/1.23 & 37.68/1.10 & 57.97/0.80 & 37.18/1.17 & 58.52/0.85\\
			\midrule
			BIOT  &3.2M   &51.18/1.18 & 44.76/2.54 & 51.36/1.12 & 38.37/1.87 & 22.61/2.62 & 38.56/2.03 &58.38/1.50 & 37.39/1.77& 58.20/1.77 \\
			LaBraM-Base&5.8M    & 52.73/1.07 & 46.98/1.88 & 52.88/1.02 & 39.76/1.38 & 23.86/2.09 & 39.74/1.11 & -- & -- & -- \\
			CBraMod&4.0M    & \underline{55.09/0.89} & \underline{50.41/1.22} & \underline{56.18/0.93} & \underline{40.91/0.97} & \underline{25.69/1.43} & \underline{41.01/1.08}& \underline{60.17/0.85} & \underline{40.58/0.91} & \underline{60.63/0.96} \\
			\midrule
			EMod &0.81M   & \textbf{62.87/0.76 }& \textbf{57.97/0.85} & \textbf{63.05/0.66 }& 
			\textbf{41.24/0.86} & \textbf{27.12/0.89 }& \textbf{42.15/0.87} & 
			\textbf{61.14/0.83} & \textbf{41.95/0.93}& \textbf{61.19/0.90}  \\
			\bottomrule
		\end{tabular}
	} 
	\caption{
		Performance comparison on three benchmark datasets. 
		Each result reports the mean/standard deviation (\%) over five runs. 
		\textbf{Bold} entries indicate the best performance; \underline{underlined} entries denote the second-best. 
	}
	\label{tab:faced_seedv_seed_results}
\end{table*}

\begin{table*}[htbp]
	\centering
	\resizebox{\textwidth}{!}{%
		\begin{tabular}{lccccccccc}
			\toprule
			& \multicolumn{3}{c}{\textbf{FACED (9 classes, 10s)}} & \multicolumn{3}{c}{\textbf{SEED-V (5 classes, 1s)}} & \multicolumn{3}{c}{\textbf{SEED (3 classes, 5s)}} \\
			\cmidrule(lr){2-4} \cmidrule(lr){5-7} \cmidrule(lr){8-10}
			\textbf{Method} & \textbf{BACC} & \textbf{Kappa} & \textbf{WF1}
			& \textbf{BACC} & \textbf{Kappa} & \textbf{WF1}
			& \textbf{BACC} & \textbf{Kappa} & \textbf{WF1} \\
			\midrule
			EMOD-Scratch & 58.99/0.94 & 53.57/1.12 & 59.37/1.06 & 
			38.47/1.00 & 24.08/1.28 & 38.30/1.10 & 
			58.35/1.04 & 37.83/1.55 & 59.00/1.19\\
			
			EMOD-Augment& 58.98/1.12 & 54.53/1.30 & 55.62/1.08 & 
			38.49/0.99 & 25.00/1.26 & 39.18/0.97 & 58.42/1.00
			&37.91/1.50&59.10/1.16\\
			EMOD-HardVA & 59.65/0.80 & 54.42/0.98 & 59.97/0.79 & 
			39.36/0.96 & 24.40/1.33 & 39.91/1.02 & 59.83/1.33
			&40.73/1.62&60.76/0.95\\
			EMOD-SoftVA & \textbf{62.87/0.76 }& \textbf{57.97/0.85} & \textbf{63.05/0.66 }& 
			\textbf{41.24/0.86} & \textbf{27.12/0.89 }& \textbf{42.15/0.87} & 
			\textbf{61.14/0.83} & \textbf{41.95/0.93}& \textbf{61.19/0.90} \\
			\bottomrule
		\end{tabular}
	} 
	\caption{Effect of pretraining components.}
	\label{tab:ablation_study_pretrain}
\end{table*}

\section{Experiments}
\subsection{Pretraining}
\subsubsection{Pretraining Datasets}
EMOD is pretrained on 8 publicly available EEG-based emotion datasets, including \textbf{DEAP}~\cite{koelstra2011deap}, \textbf{SEED-IV}~\cite{zheng2018emotionmeter}, \textbf{SEED-VII}~\cite{jiang2024seed}, \textbf{AMIGOS}~\cite{miranda2018amigos}, \textbf{MPED}~\cite{song2019mped}, \textbf{DREAMER}~\cite{katsigiannis2017dreamer}, \textbf{MAHNOB-HCI}~\cite{soleymani2011multimodal}, and \textbf{MDME} (Multimodal Dataset for Mixed Emotion Recognition)~\cite{yang2024multimodal}. Detailed information about these datasets is provided in Table~\ref{tab:pretrain_dataset}.

\subsubsection{Preprocessing}
To ensure consistency across different datasets, all the EEG signals are preprocessed using a standardized pipeline. First, a band-pass filter with a frequency range of 0.3–49 Hz is applied to remove low-frequency drifts and high-frequency noise. Second, independent component analysis (ICA) is performed to identify and remove artifacts such as ocular and muscle activity. Third, an average re-referencing technique is employed to reduce the influence of reference electrode selection. Finally, the preprocessed EEG signals are segmented into non-overlapping 10-second epochs to serve as input for pretraining, resulting in a total of 79,070 EEG samples across 8 datasets.

\subsubsection{Pretraining Settings}
The Triple-Domain Encoder employs two parallel convolutional branches—each consisting of a 2D convolutional layer with a kernel size of $1 \times 25$, 128 output channels, Group Normalization with 4 groups, and GELU activation—to extract temporal and spectral features, which are subsequently fused into unified token representations. The tokens are then processed by a 3-layer Spatial-Temporal Transformer, where each layer contains 128-dimensional embeddings, 16 attention heads, separate spatial and temporal axial attention modules, and a feed-forward network with a 512-dimensional hidden layer.

Pretraining is conducted on four NVIDIA RTX 3090 GPUs for 100 epochs, taking about 20 hours. . We used a total batch size of 288 ($9 \times 4$ per dataset) and optimize the model using AdamW with a learning rate of $5e{-4}$ and a weight decay of $1e{-4}$. The learning rate is scheduled using CosineAnnealingLR over 100 epochs with a minimum value of $1e{-7}$. Gradient clipping is applied with a maximum norm of 3.0, and the temperature parameter in the contrastive loss is set to 0.07, and the hyperparameter $d_{\text{max}}$ is chosen as 5.0.

\subsection{Evaluation}
\subsubsection{Evaluation Datasets}
To comprehensively evaluate our method, we conduct experiments on three representative EEG emotion datasets: FACED~\cite{chen2023large}, SEED-V~\cite{liu2021comparing}, and SEED~\cite{zheng2015investigating}.
\textbf{FACED} is a large-scale EEG dataset with 32-channel signals (250 Hz) from 123 subjects across nine emotion categories. Signals are segmented into 10,332 non-overlapping 10-second epochs and resampled to 200 Hz.
\textbf{SEED-V} contains 62-channel EEG signals (1000 Hz) from 16 subjects across five emotions. After resampling to 200 Hz, the data are segmented into 117,744 1-second samples.
\textbf{SEED} includes three sessions of 62-channel EEG from 15 subjects across three emotion categories. Signals are resampled to 200 Hz and segmented into 30,375 5-second samples. We follow the data split protocol proposed in~\cite{wang2024cbramod}, dividing samples into training, validation, and test sets. For SEED, we using Session 1 for training, Session 2 for validation, and Session 3 for testing.

\subsubsection{Compared approaches}
We compare EMOD with representative non-foundation approaches, including \textbf{EEGNet}~\cite{lawhern2018eegnet}, \textbf{EEGConformer}~\cite{song2022eeg}, \textbf{ContraWR}~\cite{yang2021self}, and \textbf{ST-Transformer}~\cite{song2021transformer}, as well as emotion-specific models: \textbf{TSception}~\cite{ding_tsception_2023}, \textbf{ICAPS-ReLSTM}~\cite{FAN2024105422}, and \textbf{MACTN}~\cite{SI2024108973}. Compared foundation models include \textbf{BIOT}~\cite{yang2023biot}, \textbf{LaBraM-Base}~\cite{jiang2024large}, and \textbf{CBraMod}~\cite{wang2024cbramod}. Based on the comprehensive evaluation presented in \cite{wang2024cbramod}, we adopt the results reported in that work for all overlapping methods and datasets. For methods or datasets not covered in \cite{wang2024cbramod}, we reproduce the results using their official implementations and released pretrained models to ensure fair comparability. Note that since LaBraM uses the SEED dataset during its pretraining phase, we exclude its results on SEED to ensure a fair comparison.

\subsubsection{Performance Metrics}
We adopt Balanced Accuracy (BACC), Cohen’s Kappa (Kappa), and Weighted F1-score (WF1) as evaluation metrics for multi-class emotion recognition, using Kappa for model selection during training. All reported results are presented as the mean and standard deviation over five runs with different random seeds.

\subsection{Result Analysis}
\subsubsection{Performance Comparison with Other Existing Approaches}
As shown in Table~\ref{tab:faced_seedv_seed_results}, \textbf{EMOD consistently achieves state-of-the-art performance across all the three evaluation datasets}, demonstrating its strong generalization capability in diverse EEG emotion recognition scenarios.
On the FACED dataset, EMOD notably surpasses the strongest baseline, CBraMod, by 7.78\% in BACC (62.87\% vs. 55.09\%), 7.56\% in Kappa (57.97\% vs. 50.41\%), and 6.87\% in WF1 (63.05\% vs. 56.18\%). These substantial gains indicate that EMOD more effectively captures emotion-discriminative features, especially in challenging multi-class settings.
Furthermore, EMOD maintains competitive performance on both SEED-V and SEED, underscoring its robustness to dataset heterogeneity, including differences in segment duration, channel configuration, and label distribution.

In addition to strong performance, \textbf{EMOD is remarkably lightweight}, with only 0.81 million parameters—significantly fewer than recent EEG foundation models.
It also requires much less pretraining data than large-scale models built on task-agnostic EEG and reconstruction objectives.
These results indicate that semantically guided contrastive learning on emotion-specific datasets is more effective for emotion representation learning than conventional unsupervised pretraining.

\begin{figure*}[!h]
	\centering
	\includegraphics[width=1\linewidth]{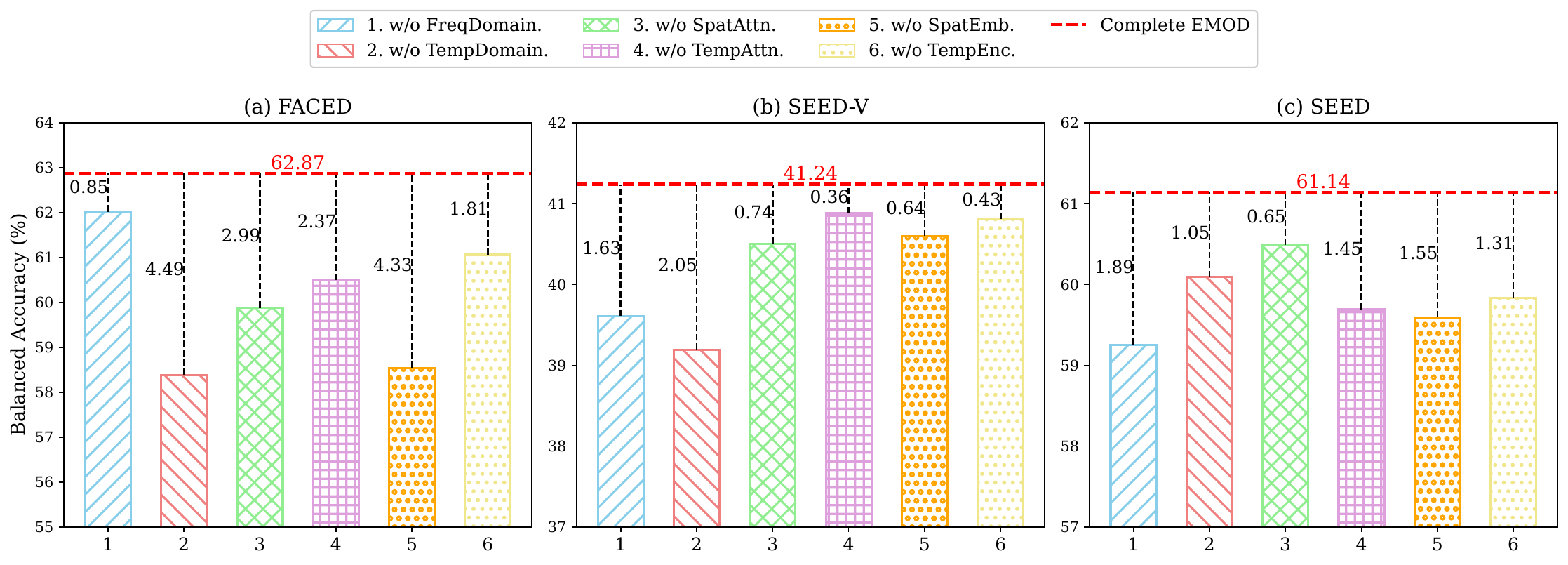}
	\caption{Ablation study on model structure. The red dashed line on top indicates the BACC of the complete EMOD model.}
	\label{fig_model_struct}
\end{figure*}

\begin{figure}[!h]
	\centering
	\includegraphics[width=1\linewidth]{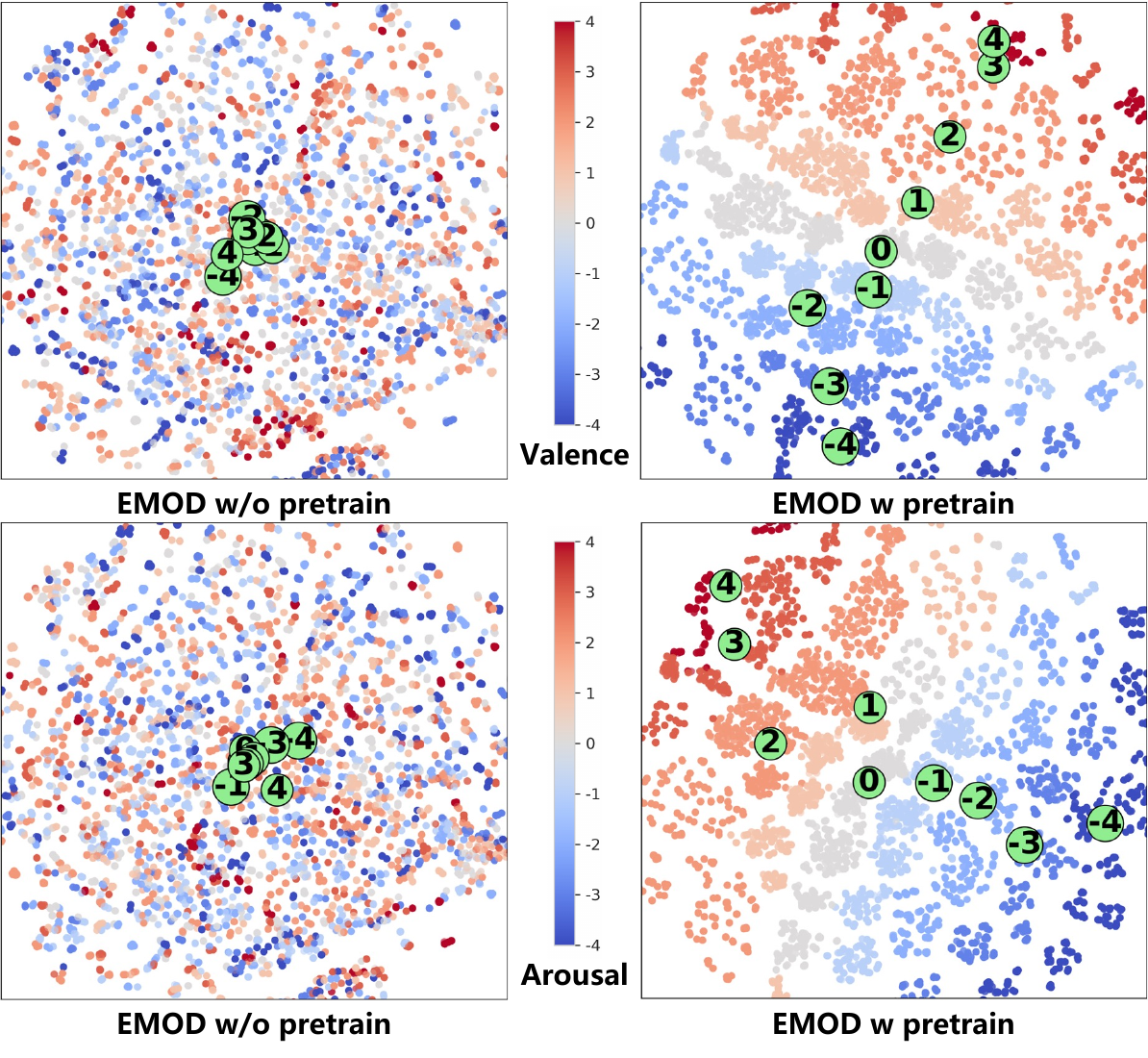}
	\caption{
		t-SNE visualization of EEG samples from the DEAP dataset. Left: EMOD w/o pretrain; Right: pretrained EMOD. Dots are color-coded by Valence (top) and Arousal (bottom), and green circled numbers indicate the centroids of discrete V-A scores (–4 to 4). Compared to the untrained model, the pretrained EMOD produces more structured and clearly separated clusters. In addition, the centroids follow a smooth emotional gradient, preserving the valence/arousal ordering in the learned representation space.
	}
	\label{fig_viz}
\end{figure}

\subsubsection{Effect of Pretraining Components}

We conduct a stepwise ablation study on the three evaluation datasets, starting from scratch and incrementally adding core components to assess their impact:
\begin{itemize}
	\item EMOD-Scratch: No pretraining; the model is trained from scratch using only downstream supervision.
	\item EMOD-Augment: Contrastive pretraining using only instance-level augmentations; no V-A guidance.
	\item EMOD-HardVA: Contrastive pretraining with V-A guided supervision, using only binary positive/negative labels without considering V-A proximity.
	\item EMOD-SoftVA: Full EMOD with soft-weighted contrastive supervision based on V-A proximity.
\end{itemize}
The results reveal three key insights as shown in Table~\ref{tab:ablation_study_pretrain}.First, EMOD-Augment offers negligible improvement over EMOD-Scratch (e.g., FACED BACC: 58.98\% vs. 58.99\%), suggesting that instance discrimination alone contributes little to emotion recognition. This is likely because augmented samples across datasets are independent and lack semantic linkage, limiting the benefits of invariance-based learning.
Second, EMOD-HardVA shows modest gains (e.g., SEED BACC: 59.83\% vs. 58.42\%), indicating that incorporating emotional semantics helps structure the latent space. However, treating all positive pairs as equally similar neglects fine-grained distinctions.
Finally, \textbf{EMOD-SoftVA achieves the best performance} (e.g., FACED BACC: 62.87\%, SEED BACC: 61.14\%), highlighting the importance of modeling graded emotional similarity. By softly-weighted contrastive pairs based on V-A proximity, EMOD learns a smoother affective representation that captures subtle emotional differences while preserving broader categorical structure.
Overall, these findings underscore that it is not merely emotional supervision, but how emotional structure is encoded—through soft, distance-aware contrastive signals—that drives better generalization.

\subsubsection{Ablation Study on Model Structure}
We conduct an ablation study to assess the role of each architectural component in EMOD by removing one module at a time and evaluating its impact on BACC across the three evaluation datasets. Tested variants include removing frequency or temporal domain tokens (w/o FreqDomain., w/o TempDomain.), disabling spatial or temporal attention (w/o SpatAttn., w/o TempAttn.), and excluding spatial embeddings or relative temporal encoding (w/o SpatEmb., w/o TempEnc.). As shown in Figure~\ref{fig_model_struct}, all components are essential—removing any leads to performance degradation. Notably, module contributions vary by dataset: removing temporal tokens causes the largest drop on FACED (4.49\%) and SEED-V (2.05\%), while SEED is more affected by frequency token removal (1.89\%). This suggests that each dataset emphasizes different signal characteristics, and EMOD’s unified design effectively captures complementary features across spatial, temporal, and spectral domains.

\subsubsection{Feature Visualization}
To illustrate how our pretraining mechanism guides the model toward emotion-aware representation learning, we visualize EEG features on the DEAP dataset using t-SNE. As shown in Figure~\ref{fig_viz}, the left panel presents features extracted from a randomly initialized model, while the right panel shows features from the pretrained EMOD model. Green circled numbers denote centroids of valence-arousal scores from –4 to 4.
In the untrained EMOD (left), features are densely entangled with no discernible structure, reflecting a lack of affective organization. In contrast, the pretrained EMOD (right) yields more structured and well-separated clusters, indicating enhanced discriminability. Moreover,  \textbf{the centroids exhibit a smooth emotional gradient, preserving the ordered progression of valence and arousal in the representation space.} These results confirm that V-A guided contrastive pretraining effectively aligns the learned representations with the continuous structure of emotional semantics.

\section{Conclusion}
In this work, we proposed EMOD, a contrastive pretraining framework for EEG-based emotion recognition. EMOD unifies heterogeneous emotion labels via Valence-Arousal guidance and employs a Triple-Domain Encoder with a Spatial-Temporal Transformer to model diverse EEG features. Extensive experiments on three benchmarks demonstrate that EMOD outperforms existing methods while using fewer parameters and less pretraining data. Ablation studies further validate the importance of our emotion-aware contrastive loss and architectural design. Our findings highlight the potential of emotionally guided pretraining for learning generalizable and affect-sensitive EEG representations.

\section{Acknowledgments}
This work was supported by STI 2030 Major Projects (2021ZD0200400), the National Natural
Science Foundation of China (No.62476240), the Fundamental Research Funds for the Central Universities (226-2025-00122) and the Key Program of the Natural Science Foundation
of Zhejiang Province, China (No. LZ24F020004). The corresponding author is Dr. Sha Zhao.

\bibliography{aaai2026}

\end{document}